\documentclass[Afour,sageh,times]{sagej}

\usepackage{moreverb,url}

\usepackage[colorlinks,bookmarksopen,bookmarksnumbered,citecolor=red,urlcolor=red]{hyperref}

\newcommand\BibTeX{{\rmfamily B\kern-.05em \textsc{i\kern-.025em b}\kern-.08em
T\kern-.1667em\lower.7ex\hbox{E}\kern-.125emX}}

\def\volumeyear{2024}
\usepackage{graphicx} 
\usepackage{amsmath} 
\usepackage{amsfonts}

\usepackage{siunitx}
\usepackage[capitalize]{cleveref}

\usepackage{array}
\usepackage{ulem}

\usepackage{xspace}

\usepackage{multirow}

\usepackage{url}

\usepackage{todonotes}
\usepackage{graphicx}

\begin{document}

\runninghead{Zhang and Yuan}

\title{PneuGelSight: Soft Robotic Vision-Based Proprioception and Tactile Sensing}

\author{Ruohan Zhang\affilnum{1}, Uksang Yoo\affilnum{2}, Yichen Li\affilnum{2}, Arpit Agarwal\affilnum{2} and Wenzhen Yuan\affilnum{1}}

\affiliation{\affilnum{1}University of Illinois, Urbana-Champaign, USA\\
\affilnum{2}Carnegie Mellon University, USA}

\corrauth{Wenzhen Yuan, University of Illinois Urbana-Champaign,
201 North Goodwin Avenue MC 258, Urbana,
Illinois, USA.}

\email{yuanwz@illinois.edu}

\begin{abstract}
Soft pneumatic robot manipulators are popular in industrial and human-interactive applications due to their compliance and flexibility. However, deploying them in real-world scenarios requires advanced sensing for tactile feedback and proprioception. 
Our work presents a novel vision-based approach for sensorizing soft robots. We demonstrate our approach on PneuGelSight, a pioneering pneumatic manipulator featuring high-resolution proprioception and tactile sensing via an embedded camera. To optimize the sensor's performance, we introduce a comprehensive pipeline that accurately simulates its optical and dynamic properties, facilitating a zero-shot knowledge transition from simulation to real-world applications. PneuGelSight and our sim-to-real pipeline provide a novel, easily implementable, and robust sensing methodology for soft robots, paving the way for the development of more advanced soft robots with enhanced sensory capabilities.
\end{abstract}


\maketitle

\begin{figure*}[tbp]
  \centering
  \includegraphics[width=\linewidth]{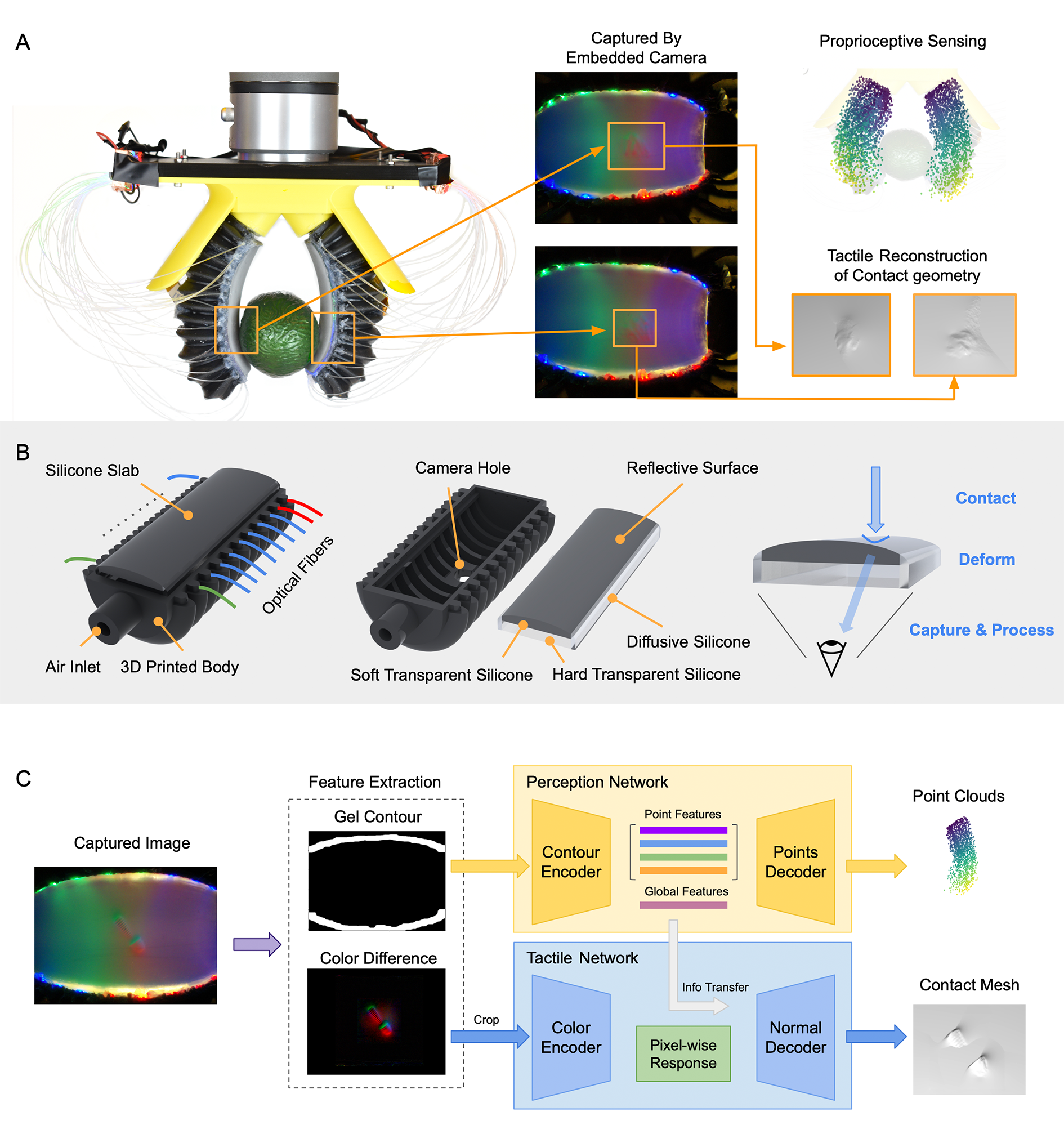}
    \caption[A, B, C]{\textbf{Design and Sensing Pipeline of PneuGelSight.} 
    \textbf{(A)} PneuGelSight gripper (left) grasping an object, with corresponding camera-captured images (middle) and high-resolution proprioceptive sensing result (top right) and tactile reconstruction of surface contact geometry (bottom right).
    \textbf{(B)} Mechanical design of the PneuGelSight sensor. The sensor integrates a deformable silicone layer, a reflective surface, and internal optical fibers for illumination. Upon contact, deformation alters the internal light pattern, which is captured by the embedded camera.
    \textbf{(C)} Data processing pipeline for sensing. A dual-branch network processes the captured image by extracting contour features for global shape reconstruction and color cues for local contact geometry.}

  \label{fig: teaser}
\end{figure*}

\section{Introduction}
Soft robots are increasingly recognized as the future of robotics due to their flexibility and intrinsic safety(\cite{rus2015design}). Their ability to comply with the external environment makes them particularly advantageous in fields such as agriculture(\cite{qiu2023tendondrivensoftroboticgripper,uppalapati2020berry}), the food industry(\cite{wang2020dual}), logistics(\cite{doi:10.1126/scirobotics.adi5908,doi:10.1126/scirobotics.adk4533}), assistive and rescue robotics(\cite{doi:10.1126/scirobotics.adi2377,doi:10.1126/scirobotics.aan3028}), and medical applications(\cite{doi:10.1126/scirobotics.adj9769,doi:10.1126/scirobotics.abq4821,doi:10.1126/scirobotics.ade2184}). However, sensing in soft robots remains a significant challenge. Unlike rigid robots, which have a limited number of joints and links, every point on a soft robot's surface can contribute to its overall shape and function, resulting in a much higher-dimensional state space. Additionally, the complex and nonlinear nature of soft materials complicates the modeling and control of soft robots. The highly deformable structure of soft robots further complicates integration with conventional rigid circuits and electronics, which are typically used in traditional sensors designed for rigid robots.

In this paper, we introduce PneuGelSight, a novel vision-based sensor designed for soft robotic fingers that combines high-dimensional proprioception and tactile sensing. The sensor is integrated into a pneumatically driven soft finger and leverages high-dimensional visual input and deep learning algorithms to model the highly nonlinear behavior of soft robots. For proprioception, we connect the 3D shape of the robotic finger to the deformation of 2D geometrical features with feature fusion technique. For tactile sensing, we apply the same principle as GelSight to measure the geometry of the contact surface. To overcome the challenge of integrating sensors within the deformable structure of a soft robotic finger, we propose an innovative design that uses optical fibers for illumination and combines data-driven methods with proprioception results. This approach ensures accurate measurement of surface geometry regardless of the overall deformation of the soft finger.
A key innovation of our method is the use of simulation. First, we employ optical simulation(\cite{agarwal2021simulationvisionbasedtactilesensors,agarwal2025modularized}) to evaluate and optimize the sensor design for tactile sensing, allowing us to identify the optimal design parameters for the soft sensor. Second, we use mechanical simulation and graphical rendering to generate large-scale deformation data and its connection to camera output to train a neural network for proprioception. This simulation-based approach provides an effective solution to the design and sensing challenges in soft robotics.

Our previous works (\cite{yoo2023toward,wang2020real}) first introduced the method of using internal vision and deep learning techniques to reconstruct the high-dimensional 3D shapes of the soft robots. The proprioception sensing capability of PneuGelSight extends this prior work by improving both the visual pattern design and algorithms. These enhancements enable the sensor to support both proprioception and tactile sensing using a single onboard camera, while also delivering significantly improved performance in shape reconstruction.
Experiments demonstrate that our system accurately senses both large-scale deformations due to actuation or environmental interaction and fine details of object surfaces in contact. This capability enables soft grippers to better understand both the global and local shape of objects, enhancing object recognition and property sensing. Our system significantly improves the ability of soft robots to perform complex grasping and manipulation tasks. The proposed methodology of using deep learning for high-dimensional modeling and simulation for sensor design can be extended to the sensing system design for other soft robots.

\section{Related Works}
\subsection{Sensing for Soft Robot}

Traditionally, soft robot sensing has relied on a combination of point-based or string-based sensing units. These sensors can be either resistive(\cite{si2023robotsweaterscalablegeneralizablecustomizable,xu2024cushsensesoftstretchablecomfortable,7353689}), capacitive(\cite{doi:10.1126/science.aac5082,ROBINSON201547}), magnetic(\cite{magnetic_nanocomposite,yan2021soft}), or based on optical(\cite{electroluminescent_devices,7487590,9426391}).  
The signals measured at each sensing unit allow for a general estimation of the robot's deformation (perception) and contact information, such as location and force (tactile sensing). However, point-based sensors face limitations in spatial resolution and require quadratic increases in resources to scale up the sensing array. In the human hand, there are approximately 17,000 tactile units (\cite{touch_sensation}), and replicating this level of sensor density while effectively processing the data remains a challenge for point-based systems. For string-based sensors, the signal is aggregated through the entire string, which causes confusion for contact area localization and limits the spatial accuracy of sensing.

Alternatively, researchers have looked into novel solutions for soft robotic sensing in recent years, including acoustic (\cite{doi:10.1177/02783649231168954}) and vision-based sensing(\cite{yoo2023toward,149155}). Acoustic sensing has been shown to work well with soft structures, achieving sensing with a simple embedded microphone. But like string sensors, the sound received at the end-effector is aggregated through the whole body, limiting the sensing resolution. Vision-based sensing, originally proposed for precise contact texture reconstruction, remains underexplored in soft robotics. 
Our previous work, \cite{yoo2023toward} , demonstrated high-resolution perception in a soft robot by leveraging a hollow structure embedded with densely distributed internal visual markers. An internal camera captured the motion of these markers during deformation, and a simple neural network was used to reconstruct the robot’s real-time shape as a dense point cloud. That study also introduced a sim-to-real pipeline in zero-shot fashion to streamline data collection for model training.

In this paper, we extend that framework to achieve high-dimensional proprioception sensing with PneuGelSight, introducing two key improvements. First, we utilize the intrinsic optical patterns of the tactile sensor as visual cues for deformation tracking, enabling simultaneous tactile and proprioceptive sensing from a single camera. Second, we redesign the shape reconstruction pipeline by incorporating a pretrained autoencoder to encode the undeformed geometry of the robot. The online sensory input is then used to estimate deformation relative to this shape prior. This new approach improves shape tracking accuracy from 8.85 mm to 5.35 mm and offers greater robustness under large deformation scenarios.

\subsection{Vision-based Tactile Sensing}

Vision-based sensing leverages embedded cameras to detect changes in optical signals related to the physical parameters being measured. Vision-based tactile sensors(\cite{yuan2017gelsight,tactip,8246881})
output data in the form of images, boasting spatial resolution to several micrometers to acquire detailed information about surface topography. A good example of this type of the sensor is GelSight(\cite{yuan2017gelsight,johnson2011microgeometry}), which measures high-resolution geometry of the objects in contact. A GelSight sensor uses an embedded camera to capture the change of light reflection from a reflective surface to infer the surface normal of the deformed sensor surface. This involves internal light sources from multiple directions, which typically are labeled with different colors. 

Various GelSight designs(\cite{zhao2023gelsight,taylor2022gelslim,tippur2023gelsight360,Mirzaee_2025}) have been developed to accommodate different robotic platforms, but they all rely on a rigid base where optical and electronic components are fixed. This structure limits deformation to a thin, flexible surface layer, typically within a few millimeters, and allows traditional vision-based tactile algorithms to operate under the assumption of a static, flat imaging plane.

In contrast, our objective is to fully sensorize a soft robot with high-resolution proprioception and tactile sensing, which fully leverages the potential of vision-based sensing principles exemplified by GelSight. While the core principle remains similar, applying it to a highly deformable robot introduces fundamental challenges. Traditional algorithms for shape and contact reconstruction fail, as they cannot accommodate non-planar deformations where the entire sensor geometry evolves dynamically. Moreover, optical signals are compromised under strain due to variations in illumination, reflection, and surface normal estimation.

Our work addresses these challenges by designing a fully deformable vision-based pneumatic gripper and a new perception pipeline that jointly enable high-fidelity proprioception and tactile sensing across the robot body. Moreover, the proposed methodology is generalizable and can be extended to a wide range of soft robotic designs. This opens the door to a new class of sensorized soft robots capable of precise interaction and perception in unstructured environments.

\section{Vision-based Sensorization}

\subsection{Pneumatic Soft Gripper with Embedded Camera}

We present the design and working principle of PneuGelSight (as shown in Figure \ref{fig: teaser}), a pneumatic soft finger that integrates an embedded camera and internal illumination to achieve both proprioception and tactile sensing. The finger's mechanical design is inspired by \cite{SoftSomatosensitiveActuators}, with a 3D-printed bellow-shaped back and an inner side made of a cast silicone slab. The silicone slab is made from a harder material and is thicker, allowing the bellow structure to elongate easily under actuation, while the silicone slab bends but resists extension. This design ensures that the finger bends inward during actuation, facilitating the grasp of the target object in conjunction with the opposing finger. 

In our design, the entire inner side of the finger serves as the contact surface. The silicone slab is made from transparent material and coated with a metallic surface layer to achieve the desired reflectivity. The camera is embedded in the back of the soft finger, as shown in Figure \ref{fig: teaser}B. The hollow structure of the finger ensures that the camera has an unobstructed view of the entire contact surface, regardless of the finger's bending angle. The deformation of the contact surface in the camera's view directly corresponds to the 3D shape of the entire finger.

The optical structure for soft robot sensing consists of three components: an embedded camera, an illumination system, and a soft reflective surface on the top of the silicone slab where contact happens. The illumination system directs light onto the reflective surface, which then scatter lights in various directions depending on the surface normals on the contact surface. The embedded camera, positioned to face the reflective surface, captures color changes caused by light reflection. These variations in light reflection are used to calculate the surface normals, which are then integrated to reconstruct the 3D geometry of the object's surface.

The design of the illumination system is critical for achieving high-quality tactile sensing and presents a significant design challenge. The goal is to create a uniformly distributed light field across the sensing surface while maintaining distinctive light reflection patterns for different surface normals. Additionally, the system must be compact and flexible to accommodate the extensive deformations of the soft finger.
To meet these requirements, we implement an array of 0.75 mm optical fibers for illumination. These fibers are adhered to the sides of the silicone slab, guiding light from a remote source to the sensing surface. At the connection points, we add a layer of translucent silicone as a diffuser to ensure that the light is evenly distributed across the surface. This design allows each optical fiber to maintain its position during finger deformation, ensuring consistent light direction and intensity. To enhance the sensing accuracy, we use different colors of light for each optical fiber, enabling precise control over the illumination angles.

\begin{figure}[tbp]
  \centering
  \includegraphics[width=\linewidth]{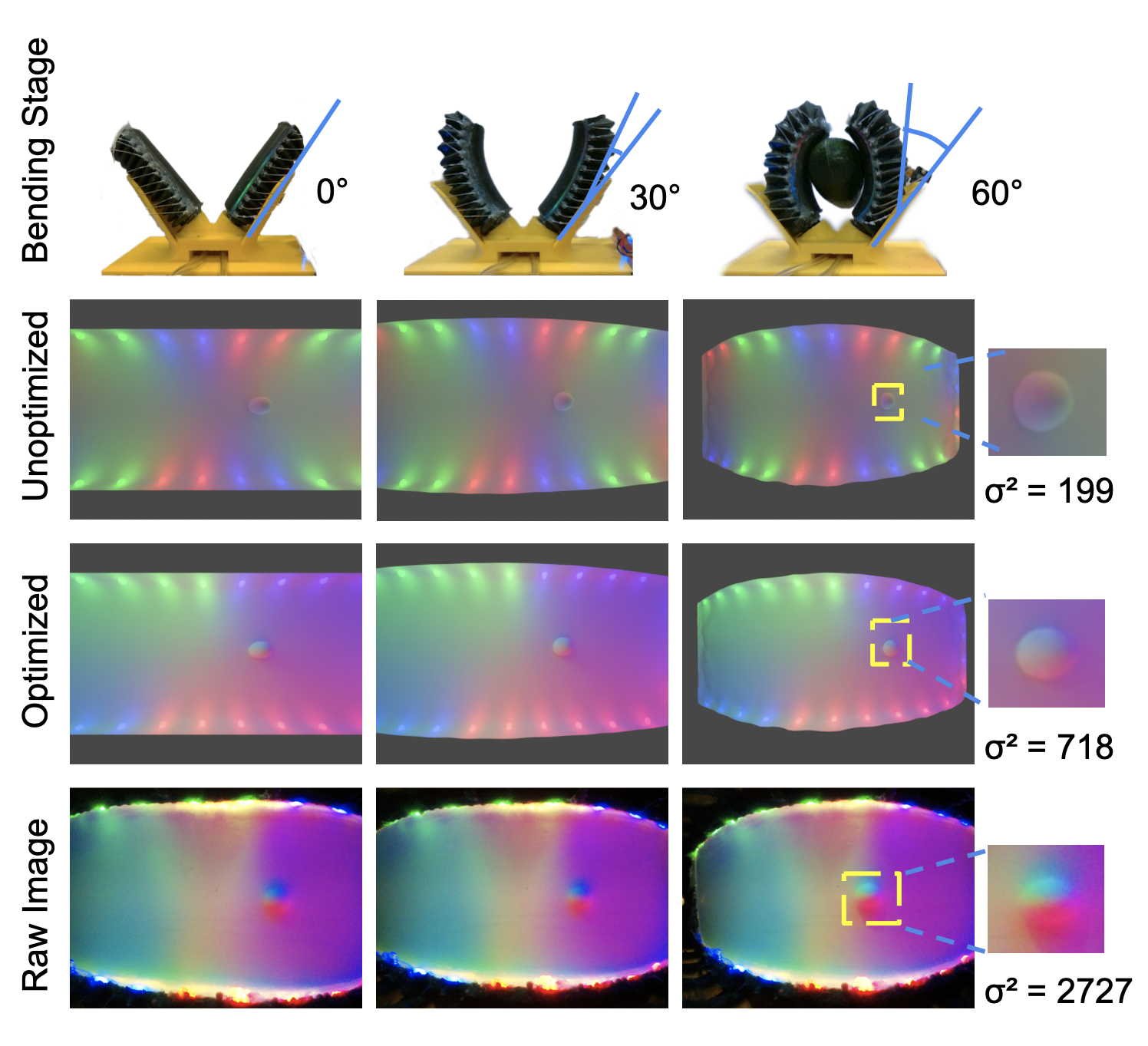}
  \caption[]{\textbf{Optimization for Optical Design.} Comparison of simulated images with various optical designs and real images under different bending angles, along with their corresponding variance scores. The design with the highest variance is selected for real-world fabrication.}
  
  \label{fig: optimize sensor}
\end{figure}

\subsection{Design and Fabrication}

Taking inspiration from the size of a human palm, we set the overall dimensions of the robot to be 110 mm long and have a semi-circular cross-section with a diameter of 55 mm. 

We use SLA 3D printing to create a two-part mold and cast the silicone slab in three steps: opaque silicone (Smooth-On, EcoFlex 00-30) as a light diffuser on the sides, crystal clear silicone (Silicone Inc., XP565) in two hardness levels by mixing silicone and curing agent at 7:1 (harder) and 14:1 (softer for better sensitivity). The soft sensing surface is then coated with semi-specular aluminum powder and protected with a final silicone layer (XP565, 14:1).

We 3D print the actuated body of PneuGelSight with a compliant and flexible silicone material (Silicone 40A). The lighting system of the robot is configured into a distinct pattern: 6 green, 7 blue, 7 red, and 4 additional blue fibers, arranged in clockwise order. We design some semi-cylindrical protrusion on the side of the robot backbone to constrain the orientation of the fibers. Each fiber is glued and inserted into the protrusion manually. After the lighting system is established, we seal the embedded camera into the camera hole and apply more silicon glue for air tightness. The camera's field of view is approximately 160 degrees (Arducam, Wide Angle Camera).

\subsection{Optimization for Optical Design}

We use simulation to assist the optical design of PneuGelSight for better sensing ability. Traditionally, developing high-resolution optical-based tactile sensors involves a repetitive trial-and-error process that demands expert knowledge. Minor changes to the optical system can significantly affect sensor performance, making it challenging to optimize the design for sensing ability. This difficulty is heightened in our soft sensing system, which must adapt to complex deformations, such as grasping objects of varying sizes, while maintaining sensing accuracy. Recently, \cite{agarwal2021simulationvisionbasedtactilesensors, agarwal2025modularized} applied physics-based rendering (PBR) techniques to simulate vision-based tactile sensors, allowing for rapid adjustments to optical elements and sensor surface properties in a simulated environment. Inspired by PBR, we simulate images to optimize color arrangements for improved sensing ability and apply the optimal design in sensor fabrication.

To transfer the optimal design from simulation to the real sensor, we must account for discrepancies between simulated and real images, as shown in Figure \ref{fig: optimize sensor}, rows 2 and 3. Overall, the color distribution aligns well, with minor edge distortions near the light source. Preliminary experiments with different color arrangements showed a consistent trend between simulated and real setups. Although not an exact match, the design quality remains consistent, enabling the transfer of the optimal design to the real sensor.

The design factor we aim to optimize in the simulation is the color arrangement, corresponding to the placement of optical fibers in the real-world sensor. We propose using color variance within the contact area as a metric to evaluate designs. The intuition behind this metric is photometric stereo(\cite{5995510}), which suggests that the sensing ability for such optical-based sensors relies on the color distinction between different surface normals—a higher distinction makes it easier for the camera to differentiate these normals. In the simulation experiment we use a sphere with radius = 5mm to indent the surface as shown in Figure \ref{fig: optimize sensor}, and the color variance within this area effectively approximates color distinction.  The variance metric $\sigma^2$ can be expressed as:

\[
\sigma^2 = \frac{1}{|A_{\text{contact}}|} \sum_{i \in A_{\text{contact}}} (C_i - \mu)^2 \tag{1}
\]

where the mean color \( \mu \) over \( A_{\text{contact}} \) is:
\[
\mu = \frac{1}{|A_{\text{contact}}|} \sum_{i \in A_{\text{contact}}} C_i \tag{2}
\]

To identify the design that best handles complex deformations, we evaluate the sensor's performance under various bending scenarios, considering different internal pressures and external contacts. For each color design, we average the performance scores across all bending scenarios and sphere-indented locations, selecting the design that maximizes the score as the final color arrangement. This arrangement is then implemented in the real-world sensor, as shown in Figure \ref{fig: optimize sensor}. In subsequent experiments, we adjust the camera's exposure rate to improve color representation.

\section{Proprioception: Sensing Robot Deformation}
The PneuGelSight sensor is designed to achieve high-resolution proprioception based on the camera's input. The proprioception, which aims to model the deformed geometry of the entire finger, can capture the fine-grained deformation caused by either actuation or collision with external objects. To achieve this, we develop a machine learning method to connect the high-resolution camera input to a high-resolution point-cloud description of the finger's shape. 

The machine learning model uses the finger's geometric features as input. Specifically, it extracts the front surface contour from a binary image and outputs a feature describing the finger's detailed deformation based on the original point-cloud shape.
To improve the efficiency of data collection, we develop a physics-based simulation framework to collect the camera's data under different robot deformation cases, and then perform a zero-shot transfer to the real robot.
Our new method includes a PointNet(\cite{qi2017pointnetdeeplearningpoint}) autoencoder to address the new challenge of the geometrical feature as input since the complicated nature of the sensor design contained our choice of customizing the geometrical markers to extract from the finger. Unlike the previous method, which directly predicts the finger's shape, our new model learns a deformation feature that is applied to the original shape and efficiently captures high-dimensional deformation from simple input features by incorporating shape priors.

\begin{figure}[htbp]
  \centering
  \includegraphics[width=\linewidth]{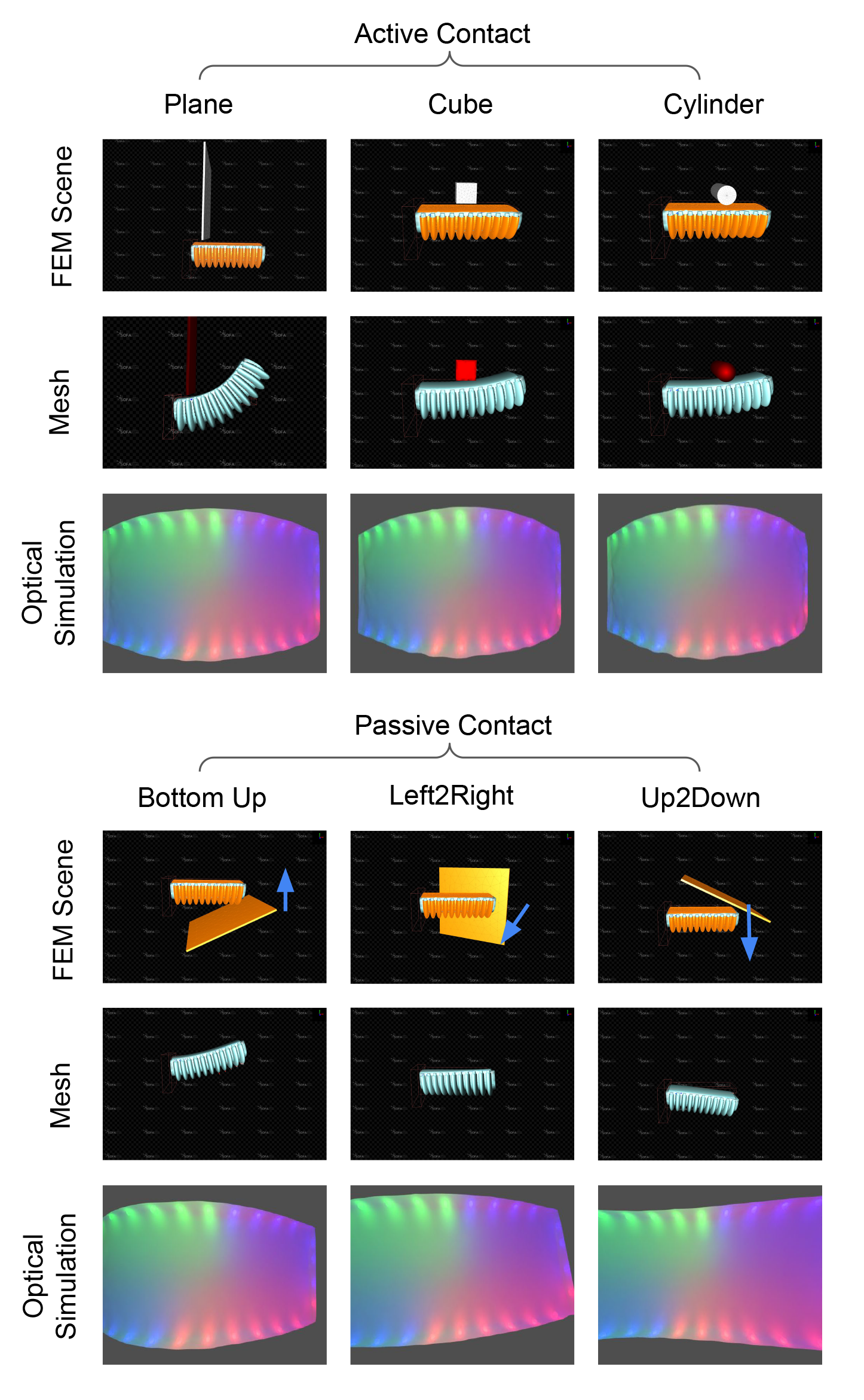}
  \caption[]{Overview of FEM scenes in the dataset, with the blue arrow indicating the movement direction of the external object.}
  \label{fig: simulated dataset}
\end{figure}

\subsection{Data Collection} Our primary focus with dynamic FEM simulations is on the robot’s interactions with objects, during which it undergoes deformation either actively or passively. We record all these states during each trial and export detailed mesh data for further analysis.

Two factors are randomized for augmentation during the simulation: the pressure of pneumatic actuation and contact conditions with an external object. We illustrate 6 selected bending scenarios in Figure \ref{fig: simulated dataset} as a demonstration. Sequentially, the actuated robot interacts with a series of fixed objects—a wall, a cube, and a cylinder—and a moving plane, approaching from various directions including bottom-up, from the side, and top-down. 
For each bending case, we record the dynamic process until the robot reaches a steady state, and save all the mesh data to build the FEM dataset.
In total, we simulate 26 different scenes and produce 3,000 images corresponding to the FEM dataset. To reduce the gap between real and simulated data distributions, we augment the simulated images during training, improving the model’s ability to generalize to real-world conditions and enhancing its robustness to variations not present in the simulated dataset.

\begin{figure*}[htbp]
  \centering
  \includegraphics[width=\linewidth]{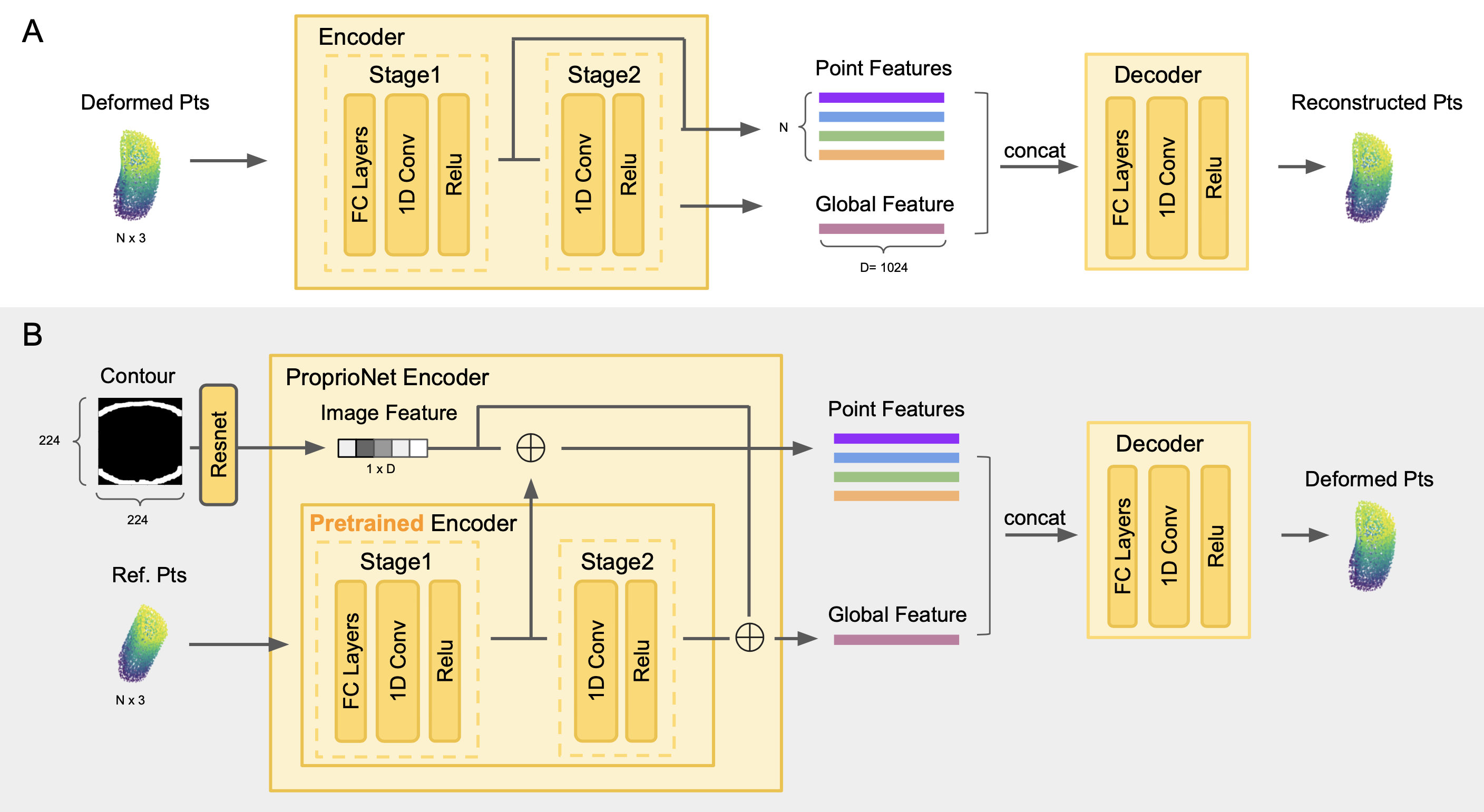}
\caption[A, B]{\textbf{Proprioception Pipeline.} \textbf{(A)} Pre-training the auto-encoder to reconstruct robot shape points cloud. \textbf{(B)} The architecture of proprioception network (ProprioNet). The feature extracted from the shape reference is modified by image feature in the latent space, and then mapped to the deformed point clouds.}
  \label{fig: proprioceptive network}
\end{figure*}

\begin{figure*}[htbp]
  \centering
  \includegraphics[width=\linewidth]{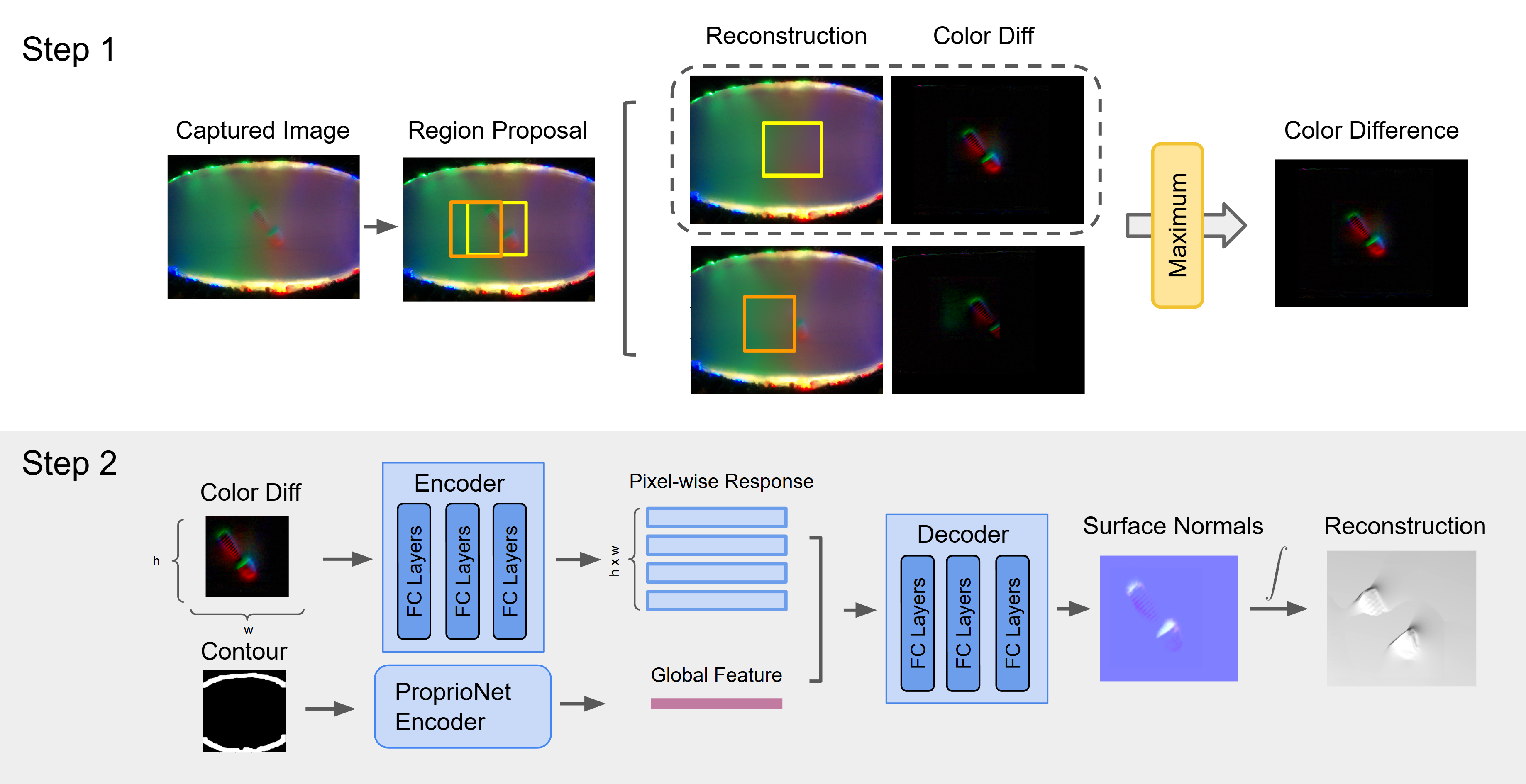}
  \caption[A, B]{\textbf{Overview of the tactile sensing network}. Step 1, The raw image captured by the embedded camera is processed through region proposal and background subtraction to extract a color difference map highlighting the contact area. Step 2, The identified region is then fed into the 3D mesh reconstruction network to estimate the contact surface geometry. }
  \label{fig: tactile network}
\end{figure*}

\subsection{Network Architecture}
An overview of the proprioception model architecture is presented in Figure\ref{fig: proprioceptive network}. First, we pre-train a feature extractor for robot shapes. It is designed to capture both low-level and high-level features from dense point cloud data, reducing noise introduced by using high-dimensional representations. The network architecture is adapted from PointNet (\cite{qi2017pointnetdeeplearningpoint}), with modifications to the depth of the encoder and decoder. In each training step, we sample $N=4096$ points from a single robot mesh within the FEM dataset and try to reconstruct the same point clouds with decoder's output. The loss for this stage $L_{\text{recon}}$ is defined as:
\begin{align*}
  &L_{\text{recon}} = \text{CD}(p, p_{\text{recon}}) \\
&= \frac{1}{|p|} \sum_{x \in p} \min_{y \in p_{\text{recon}}} \| x - y \|^2 + \frac{1}{|p_{\text{recon}}|} \sum_{y \in p_{\text{recon}}} \min_{x \in p} \| y - x \|^2
\tag{3}  
\end{align*}

Where $p$ stands for point clouds $p \in R^{N\times 3}$, and CD indicates standard chamfer distance. This stage is noted as the `pre-train' stage in the following text. 

After the pre-training stage, we train a conditional PointNet-based network to predict point cloud deformation using contour data derived from the binarized captured image.
The Encoder and Decoder modules in Figure \ref{fig: proprioceptive network}B are adapted from the pre-training stage, with weights of the Encoder being frozen during subsequent training. During the second training process, We sample point clouds from both the undeformed and deformed mesh with a density of $N$ and extract the corresponding contact gel contour as binary mask. Based on a Resnet module (\cite{he2016deep}), our image encoder encodes the binary mask into a latent visual feature vector. This feature is spatially repeated and added to the reference point and global features through element-wise summation. The reference features are extracted from the undeformed mesh noted as shape ref.
In summary, the second stage can be viewed as a conditional regression problem, integrating visual inspection into the trained auto-encoder to guide deformation prediction within the same 3D space.

During the second training stage, the loss function integrates the Mean Squared Error (MSE) between the current global feature $g$ predicted by the multi-modal ProprioNet encoder, and the global feature $g_{\text{pre-trained}}$, extracted by the auto-encoder. Additionally, we keep Chamfer Distance as part of the loss calculation to enforce low-level similarity. The loss function for the second stage is then formalized as follows:

\[L = L_{\text{recon}} + \frac{1}{N} \|g - g_{\text{pre-trained}}\|^2 \tag{4}\]

\section{Tactile Sensing: Surface Textures Reconstruction}


3D reconstruction of the contact surface provides the most precise tactile feedback available for all tactile sensors. It helps in understanding the surface property of the object, facilitating the estimation and manipulation process.
While small, rigid GelSight sensors can easily achieve surface reconstruction using a lookup table, achieving similar results algorithmically with a deformable GelSight sensor under complex lighting conditions remains a challenge for the entire community(\cite{she2020exoskeleton,ma2024gellinkcompactmultiphalanxfinger}). To address this, we propose a novel pipeline based on mathematical estimation and machine learning, as shown in Figure \ref{fig: tactile network}.

Our pipeline consists of three modules: the region proposal module, the background estimation module, and the 3D reconstruction module. The first two modules are combined as a 'pre-selection' process, predicting the most probable contact area for each captured image. During this process, the region proposal module first generates a fixed number of bounding boxes as contact proposals that cover the whole sensing area (36 boxes in the experiment), and the background estimation module then predicts a background image for each of these proposals. In those estimated background images $I_{background}$, the contact feature is removed and possible color under the same deformation state fills the contact regions. For each of those $(I_{image}, I_{background})$ pairs, the color difference score $\Delta C$ is calculated as  
\[\Delta C = \sum|I_{image} - I_{background}| \tag{5}\]
where $I_{image}, \, I_{background} \in R^{H \times W \times 3}$, and $H, W$ are the height and width of captured images, respectively. The proposal with the highest $\Delta C$ is selected as the most plausible contact area and sent to the reconstruction module.

After the pre-selection stage, the reconstruction module predicts the surface normals $(N_x, N_y, N_z)$ for all the pixels in the contact region, and Poisson integration is applied to reconstruct the 3D mesh. We demonstrate the whole tactile reconstruction process in Figure \ref{fig: tactile network}.

\subsection{Region Proposal and Background Estimation Module}
The pre-selection process identifies the most probable contact region within the captured image. Unlike traditional U-Net-based (\cite{ronneberger2015unetconvolutionalnetworksbiomedical}) segmentation methods, our module requires no additional training and offers improved generalization across different bending scenarios. In our experiment, the region proposal module generates uniformly distributed bounding boxes over the 2D image space, followed by background image estimation for each proposal.

Before diving into the details of background estimation, we first build a physical model for the lighting process. For a spotlight $j$, it's light intensity value $I_{i, j}$ at pixel position $i$ can be expressed as the follows:

\begin{align*}
& I_{i, j} = A_{i,j} \times I_{j} = \frac{1}{d_{ij}^2} \cos(\theta_{ij}) \cos(\phi_{ij})^\alpha \times I_{j} \tag{6}\\
& I_{i} = \sum_{j=1}^{24} I_{i, j} \tag{7} \\
\end{align*}

Where $I_{i}$ stands for the light intensity at pixel $i$. In this function \(d_{ij}\) represents the distance between the pixel \(i\) and light \(j\), \(\theta_{ij}\) is the angle between the pixel's normal and the light direction (calculated as \(\arctan\left(\frac{d_{ij}}{z}\right)\), with \(z\) denoting the thickness of the contact gel), and \(\phi_{ij}\) is the angular displacement in the 2D plane between the spotlight-like \(j\)-th light and the \(i\)-th pixel. This expression combines the inverse square law to adjust for distance, and a directional cosine component, modified by an exponent \(\alpha\), to account for the alignment and angular displacement between the pixel and light source.

Assume a color space where the color value is proportional to light intensity \( I \), and the relationship is expressed as \( b = Ax \), where:
\begin{itemize}
\item \( b \in \mathbb{R}^{n \times 3} \) represents the linearized color values
\item  \( A \in \mathbb{R}^{n \times 24} \) is a coefficient matrix calculated by (6)
\item \( x \in \mathbb{R}^{24 \times 3} \) denotes the light intensity, which is scaled by a constant factor \( \alpha \) such that \( x = \alpha I \)
\end{itemize}
We sample \( n \) points near the contact area to obtain color values, and estimate the light intensity \( x \) from the coefficient matrix \( A \) and the measured color values \( b \) by solving the inverse problem, \( x = A^{-1}b \). Once \( x \) is estimated, the background color within the contact region can be computed through forward estimation, \( b' = A'x \), where \( A' \) is a new coefficient matrix that depends on the pixel location.

To change the representation of color to a space where the value is linearly related to the intensity, we can apply the following color space transformation:

\begin{align*}
C_{Linear} &= 
\begin{cases} 
\frac{C_{RGB}}{12.92} & \text{if } C_{RGB} \leq \tau_1 \\
\left(\frac{C_{RGB} + 0.055}{1.055}\right)^{2.4} & \text{if } C_{RGB} > \tau_1
\end{cases}\tag{8}
\\
C_{RGB} &= 
\begin{cases} 
C_{Linear} \times 12.92 & \text{if } C_{Linear} \leq \tau_2 \\
1.055 \times C_{Linear}^{(\frac{1}{2.4})} - 0.055 & \text{if } C_{Linear} > \tau_2
\end{cases}\tag{9}
\end{align*}
with $\tau_1, \tau_2$ being the threshold for illumination.

\begin{figure*}[htbp]
  \centering
  \includegraphics[width=\linewidth]{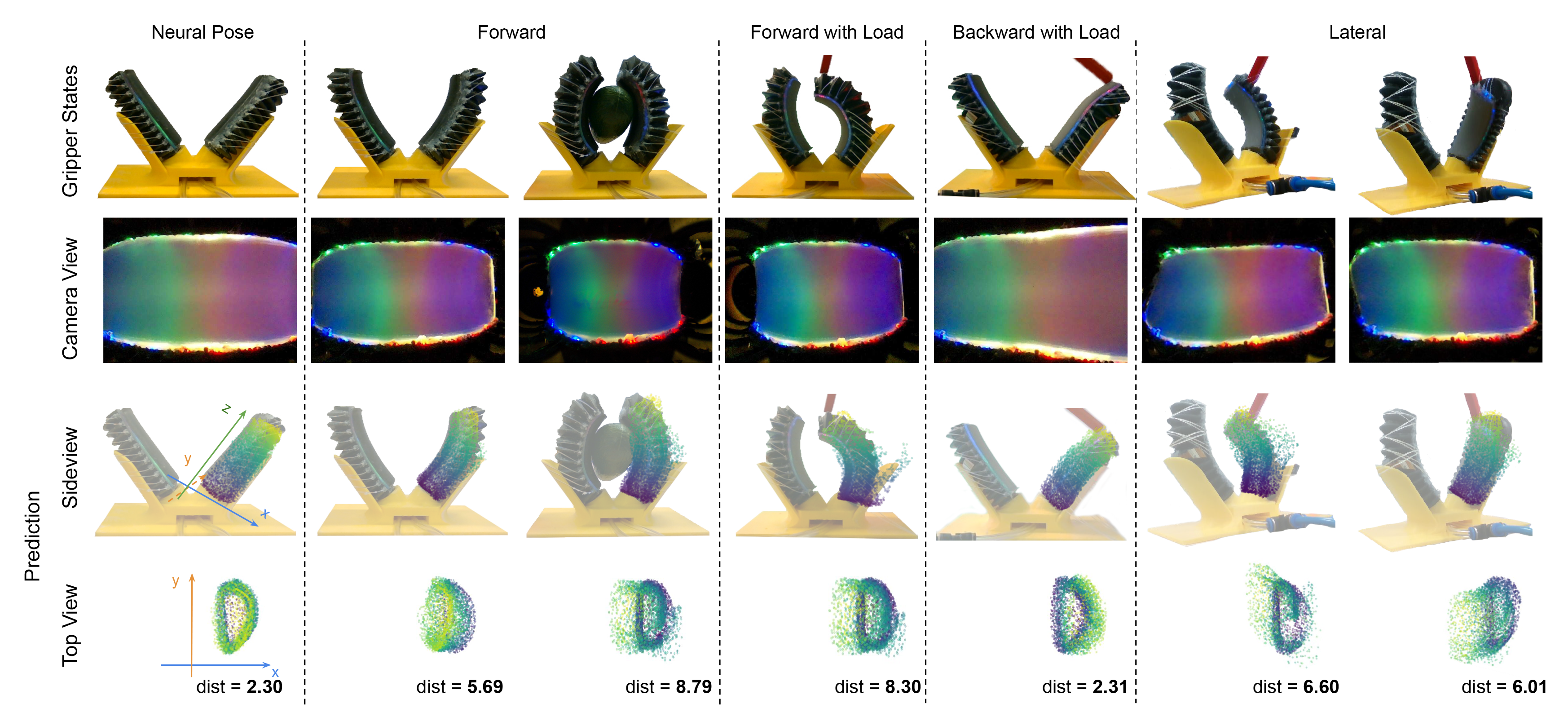}
  \caption[]{\textbf{Result of high-resolution proprioception.} We categorize the gripper’s deformation into five representative scenarios and present the corresponding external view (top row), internal camera view (middle row), and predicted point clouds (bottom row) from two viewpoints. The point clouds are color-coded based on the Z-axis values.  The Chamfer Distances between the predicted and ground truth point clouds are reported for each case to quantify reconstruction accuracy.
  }
  \label{fig: proprioceptive demo 1}
\end{figure*}

\begin{figure}[htbp]
  \centering
  \includegraphics[width=\linewidth]{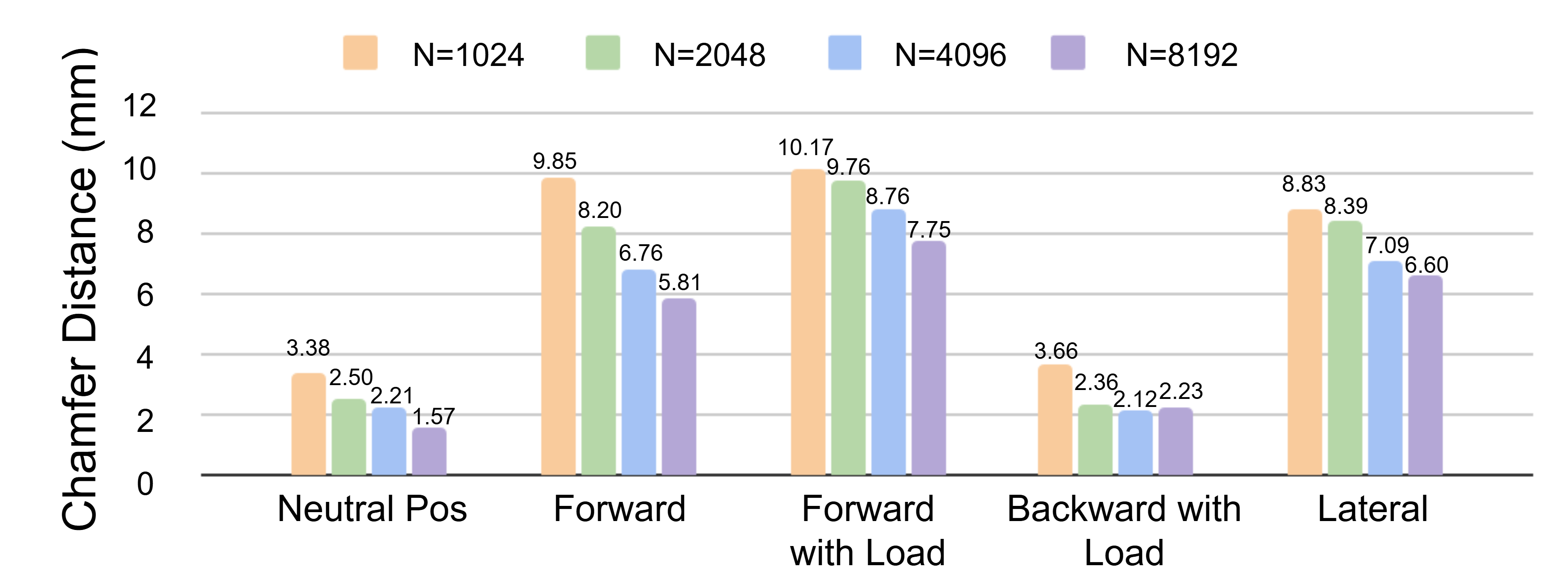}
  \caption[]{Effect of point cloud density on proprioception accuracy.}
  \label{fig: proprioceptive demo 2}
\end{figure}

\subsection{Network Architecture}
We construct a dataset of $(I_{image}, N)$ pairs to train the reconstruction network, with $N = (N_x, N_y, N_z)$ standing for the surface normals. To create the dataset, we indent the real PneuGelSight sensor with a 5mm radius sphere at various locations and manually labeled the surface normal. During the training process, the region proposal module is disabled since the ground truth contact locations are known. We use the background reconstruction module to calculate the color difference and use it as the main input to the reconstruction network. Considering that MLP doesn't require spatial information, We rearrange those data accordingly to be of shape $R^{m\times 29}$, where $m = 2000$ is the number of pixels used in one iteration, and 29 means that for each pixel, all color value within a $3 \time 3 $ range ($\in R^{1\times 27}$) as well as its (x, y) position ($\in R^{1 \times 2}$)are considered when doing the reconstruction. To gather enough training samples of $m$, we use all pixels within the contact region as well as some points outside the contact range, with their corresponding surface normal set to be (0, 0, 1).

The reconstruction network is a conditional MLP with an Encoder-Decoder structure. A 3-layer MLP encoder extracts color features and integrates global features from the pre-trained ProprioNet, as mentioned earlier. The decoder, also a 3-layer MLP, reconstructs surface normals using the fused features. The network is trained using a simple MSE loss between the predicted normals and the ground truth normals. Figure \ref{fig: tactile network} illustrates our complete pipeline for 3D reconstruction from a single image. To our knowledge, this is the first successful method for reconstructing the contact geometry of a fully deformable vision-based tactile sensor.

\section{Experimental Evaluation of Sensing Ability}

\subsection{High-Resolution Proprioception}

\begin{figure*}[htbp]
  \centering
  \includegraphics[width=.98\linewidth]{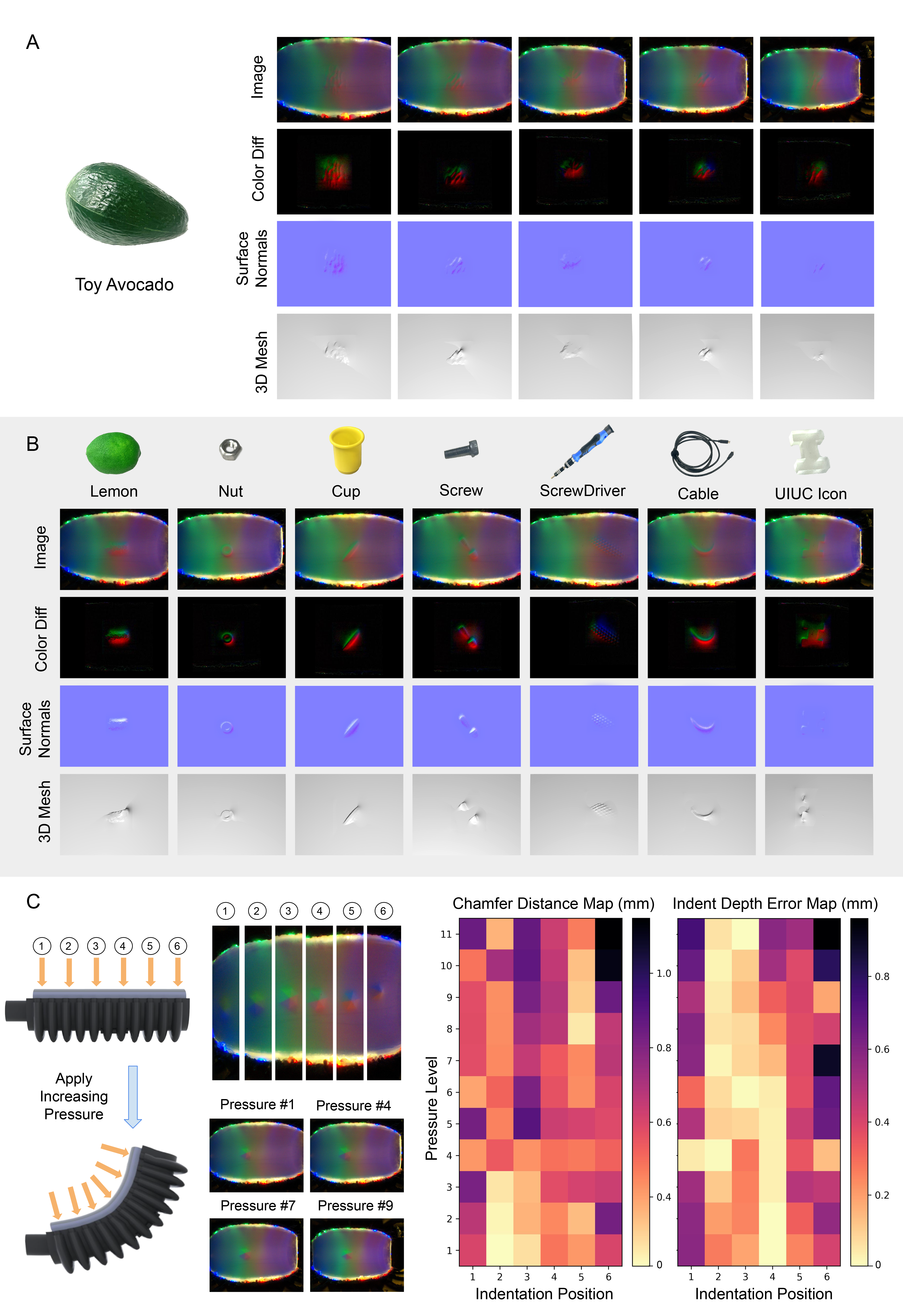}
   \caption[]{\textbf{Tactile sensing results.} \textbf{(A)} Tactile Imprints of an avocado under different bending angles, along with the corresponding color differences, predicted surface normals, and reconstructed 3D meshes. \textbf{(B)} Imprints from different objects. \textbf{(C)} Quantitative evaluation of the sensor’s tactile performance using two metrics: Chamfer distance and maximum indentation depth error. Experiments were conducted across 11 pressure levels and 6 indentation positions along the sensor's longitudinal axis.}
  \label{fig: tactile 1}
\end{figure*}

\begin{figure}[htbp]
  \centering
  \includegraphics[width=\linewidth]{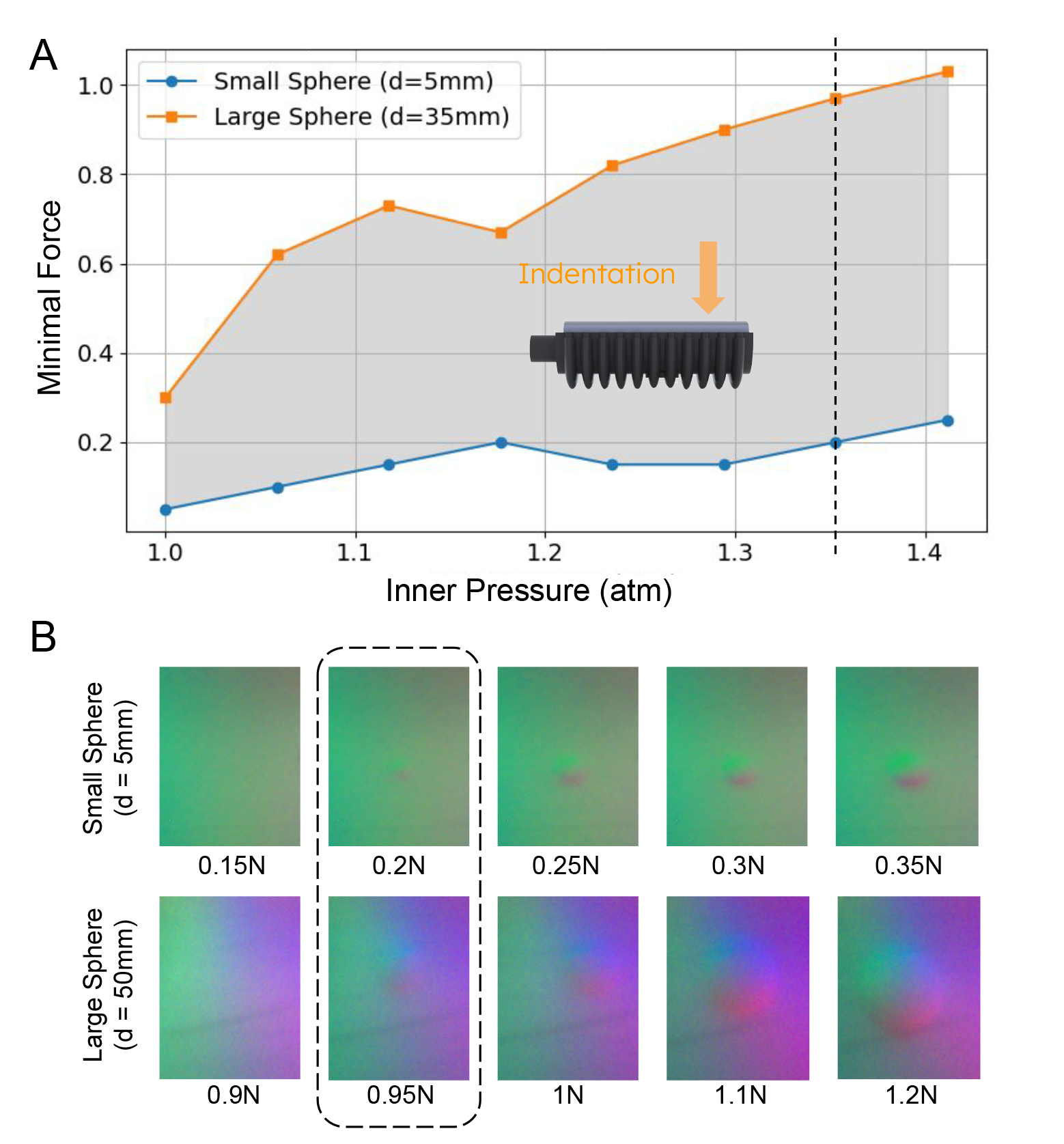}
   \caption[]{\textbf{(A)} Relationship between minimum detectable force and internal pressure for two indenters: a small sphere (5mm diameter) and a large sphere (35mm diameter).  \textbf{(B)} Cropped image illustrating the sensor’s response to varying applied forces at an inner pressure of 1.35 atm. }
  \label{fig: tactile 2}
\end{figure}

A demonstration of the proprioception result is shown in Figure \ref{fig: proprioceptive demo 1}. We quantify the results by measuring the standard chamfer distance in mm between the ground truth captured by RGB-D camera and the predicted points, where lower values indicate better accuracy. The metric is calculated through the whole actuation process, and for simplicity, we only visualize the most significant scene in Figure \ref{fig: proprioceptive demo 1}.

We select five general test cases, two of which are neutral pose and natural bending to represent typical scenarios during gripper operation, and others are bending forward, backward and lateral bending with external load. We present both side view (aligned with external image) and top views to help better interpret the point clouds. As visualized in Figure 7, our model achieves Chamfer Distances of 2.21, 6.76, 8.76, 2.12, and 7.09 mm for each test case, using a point cloud density of N=4096. The mean Chamfer Distance across all collected cases is 5.35 mm. Notably, even in the edge case of lateral bending—which is mechanically constrained and potentially harmful in real-world scenarios—the model demonstrates reasonable accuracy in reconstructing the deformation.

We further analyze the effect of point cloud density on inference performance. As expected, increasing density leads to improved accuracy but also higher computational cost and longer inference time on limited computational resources. To evaluate the feasibility of real-time deployment, we benchmark the model on an NVIDIA RTX 4070 GPU. In this setting, the model achieves inference times below 0.05 seconds even at the highest resolution (N=8192), demonstrating its capability to deliver high-precision proprioceptive feedback suitable for real-time control tasks.

\subsection{Vision-based Tactile Sensing}

We present both qualitative and quantitative tactile sensing results in Figure \ref{fig: tactile 1}.  It is evident that external contact geometry generates a strong optical signal in the tactile image, and with our proposed pipeline, we can distinguish between different surface normals of the contact object by analyzing color differences and reconstruct a 3D mesh of the texture.

Our method is robust across different actuation states of the finger. Figure \ref{fig: tactile 1}A shows tactile sensing results for various bending angles. Although larger bending angles alter the light distribution and reduce sensing ability, our algorithm ensures reliable reconstruction and keeps tactile sensing largely unaffected at all stages.

In Figure \ref{fig: tactile 1}B we visualize sensing results with common objects pressed onto the surface. For objects with pronounced textures and significant height variations, such as nuts, screws, and screwdrivers, our system reconstructs their textures with high fidelity. Even for objects with finer details, like a lemon, the rough surface is captured. A key feature of our sensor is its varying sensing ability across the large surface, with reduced resolution near the tip. When pressing a UIUC icon, the system disregards the rightmost area, as fine details are harder to perceive in that region.

To better quantify the tactile reconstruction performance of the sensor across varying resolutions and spatial positions, we introduce two key metrics: (1) the Chamfer distance between the geometry of the ground truth indenter and the reconstructed mesh, which assesses the ability of the system to resolve fine textures; and (2) the maximum depth error in the indentation, which reflects accuracy in capturing contact depth. For this evaluation, we 3D print a small six-faced pyramid indenter (8mm diameter, 2mm height). We conducted experiments at six distinct indentation locations along the longitudinal axis of the sensor and at eleven pressure levels, ranging from neutral pose (1 atm) to standard operating pressure (1.35 atm). The resulting error distributions are visualized in a grid-form error map, as visualized in \ref{fig: tactile 1}C.

On average, the best Chamfer distance score in all tests is 0.18 mm, with the lowest errors observed in the central region of the sensor at lower pressure levels. The maximum depth error remains below 0.2 mm in the central region for all bending angles, demonstrating high precision in estimating indentation depth. While the sensor’s sensitivity decreases slightly when indentations occur near the border regions or under higher internal pressures, it still produces reasonable and reliable reconstructions across all tested conditions. These results confirm improved reconstruction fidelity in areas with better visual coverage and moderate actuation, highlighting the robustness and generalizability of our proposed sensing system.

\begin{figure*}[htbp]
  \centering
  \includegraphics[width=1.0\linewidth]{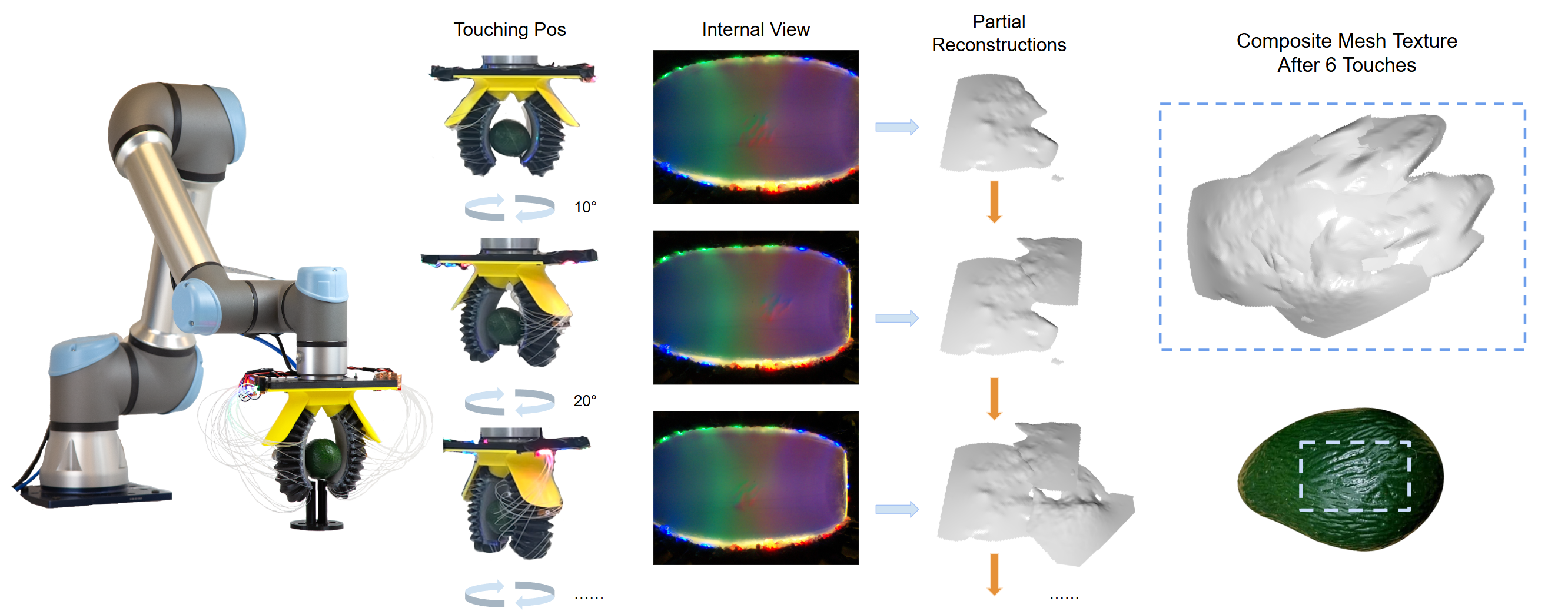}
  \caption[]{Pipeline for estimating an object's shape and texture through multiple touches. We mount our gripper on a UR5e robot arm, rotating it to generate different touching poses. The reconstructed surface texture is then stitched together for the final prediction.}
  \label{fig: cool demo}
\end{figure*}

\subsection{Tactile Sensitivity}

We define tactile sensitivity as the system's ability to resolve small contact forces and detect surface textures, which is important but hardly evaluated in current vision-based sensors. To evaluate this aspect, we employ two spherical indenters: a small sphere with a diameter of 5mm and a large sphere with a diameter of 35mm. These represent the lower and upper bounds of contact area sizes that our sensor can reliably detect. Consequently, we assume that the sensor’s sensitivity for other objects falls within the performance range established by these two reference cases.

We conducted experiments using a UR5e robot arm equipped with those two indenters, along with a force-torque sensor (Nordo Robotics, NRS-6050-D80) mounted between the end-effector and the indenter. The arm was programmed to press against a PneuGelSight sensor with constant force, while the sensor was constrained to a flat shape to maintain consistent lighting conditions across different pressure levels. In Figure \ref{fig: tactile 2}B, we present a series of x-axis cropped images, each corresponding to different normal forces, with the internal pressure set at 1.35 atm. The indenter’s surface texture first becomes visible in the second image of the sequence, which we define as the detection threshold. The corresponding force value is then taken as the minimum detectable force.

In Figure \ref{fig: tactile 2}A, we plot the relationship between minimum detectable force and internal pressure for both indenters. As internal pressure increases, the minimum force required for reliable detection also rises. Based on this data, the sensor demonstrates a sensitivity range approximately between 0.2N and 0.95N at 1.35atm, reflecting its ability to resolve subtle contact forces across varying object sizes.

The sensor's tactile sensitivity is primarily influenced by the softness of the contact layer, which can be controlled by adjusting the silicone-to-curing agent ratio during fabrication. Softer layers enhance the sensor's ability to detect finer details and smaller forces, making it suitable for delicate tasks. However, for tasks that require higher force, such as twisting an apple off a branch, increasing the hardness of the contact layer provides better structural support and stability under load.

This trade-off between force and precision has been a focus of recent research(\cite{doi:10.1177/0278364920910465}), which highlights the balance between maintaining tactile sensitivity and enabling the sensor to withstand greater forces. In our gripper design, this balance is critical. A stronger, more rigid backbone further enhances the sensor's ability to handle heavier tasks without sacrificing the necessary tactile feedback required for precision tasks.


\section{Evaluation of PneuGelSight in Real World Applications}

The human hand can grasp an object, rotate it, and quickly form an estimation of its physical properties with a few touches. This process relies on the integration of proprioception to evaluate the object's shape and tactile feedback to sense subtle surface features. In the agricultural world, such a combination of sensing is emphasized as farmers rely on the size and texture of fruit to predict its ripeness. Traditional robotic end effectors are often rigid and lack advanced sensing capabilities, which can lead to damage when handling fruit. In contrast, our PneuGelSight sensor is well-suited for detecting these features, providing a gentler and more accurate approach to fruit handling. In the following experiment, we build a real-world scenario of estimating an avocado's shape and texture with multiple touches.  We visualize the whole process in Figure \ref{fig: cool demo}.

We mount our soft gripper on a UR5e robot arm, and the avocado is lying flat on the table surface without any prior knowledge of its shape. At each time step, the last joint of the robot arm rotates at a predefined angle, and the gripper grips the avocado while capturing an image from inside the gripper. Each image provides an estimated bending shape of the gripper and a detailed contact texture reconstruction. Combining these results, we achieve a comprehensive shape and texture reconstruction of the avocado, as shown in the sensing result column of Figure \ref{fig: cool demo}. The reconstructed meshes are then stitched together for the final reconstruction. The predicted surface shows strong alignment with each other, and although the sensing area is currently focused on the middle part of the avocado due to the gripper's partial use of its capacity, the overall sensing area offers significant potential. 

\section{Discussion: Design Considerations for Illumination}
We selected optical fibers as the primary method for illumination due to their unique advantages in the context of soft robotic design. Optical fibers are lightweight and flexible, allowing them to be integrated directly into the robot body without the need for mounting rigid PCBs or discrete light sources onto the soft structure. This minimizes added mass and preserves the robot’s natural compliance and mobility. Additionally, optical fibers can produce strong, directional illumination, which is critical for achieving clear deformation-induced visual cues needed for vision-based sensing. 

While embedding optical fibers introduces some structural elements into the robot, careful routing and placement ensure that these components do not significantly impair motion or introduce unwanted stiffness. Future designs could further optimize this integration by embedding fibers within the robot’s backbone to reduce the number of external fibers. Moreover, the use of optical fibers provides a modular foundation for alternative lighting strategies, such as flexible light strings or multi-spectral sources, which may be incorporated in future iterations without major mechanical redesigns.

\section{Summary}

We propose integrating a vision-based sensing system into a soft robot design, offering two key contributions: PneuGelSight—a soft gripper equipped with high-resolution proprioception and tactile sensing via an embedded camera, and a physically accurate simulation pipeline for dynamic and optical behaviors of soft manipulators. Our PneuGelSight sensor, combining precise point cloud proprioception data with high-resolution tactile feedback, overcomes soft robotic sensing challenges with a single integrated camera. Our simulation pipeline facilitates the design, evaluation, and optimization of such soft manipulators to achieve optimal performance. It also enables zero-shot transfer from simulation to real-world scenarios. On the algorithm side, we propose a pair of interconnected neural networks. Each network is designed to process one sensing modality from the same input, and we enable information exchange to enhance overall sensing capability. To the best of our knowledge, we are the first to unify different high-resolution sensing modalities for robotic sensing,  resulting in a compact and efficient sensing pipeline.

Compared to existing sensing approaches, our solution is more straightforward to implement while delivering high-fidelity results in both proprioception and tactile sensing.  Both qualitative and quantitative evaluations confirm the sensor’s reliability, with low reconstruction and alignment errors.

\begin{acks}
We extend our heartfelt thanks to Amin Mirzaee for his invaluable assistance and expertise in the fabrication of soft robots, and Yuchen Mo for his exceptional guidance on the neural networks design.
\end{acks}

\section{Statements and declarations}
\subsection{Funding statement}
The author(s) disclosed receipt of the following financial support for the research, authorship, and/or publication of this article: This work was supported by the National Science Foundation [grant number 2348839].
\subsection{Data availability}
All the design files can be find in this repository: 

\url{https://drive.google.com/drive/folders/1_cb1CJaEQN_CRJUpdzZMATWkf7SGQSAK?usp=share_link}

\bibliographystyle{SageH}
\bibliography{ccs-sample}

\clearpage

\newpage
\appendix
\renewcommand{\thesection}{\Alph{section}}

\section{Fabrication}

The PneuGelSight sensor consists of three parts: a silicone slab, a backbone, and a lighting system with optical fibers and internal camera. To fabricate the silicone slab, we first design a two-part mold using 3D printing. After assembling and sealing the mold to prevent leakage, we sequentially pour in diffusive, hard, and soft silicones, allowing each layer to cure fully before adding the next. Once cured, we remove the top mold, apply aluminum powder as a semi-specular layer, and add a protective silicone layer. Simultaneously, we 3D print the backbone using 40A silicone and cure it with ultraviolet light.

After the silicone slab and backbone are prepared, we seal the silicone to the backbone. A small intrusion within the backbone walls indicates the correct sealing position. The height of the hard silicone is slightly larger than the sealed depth, leaving a portion of the side area uncoated and exposed to external light. This exposure enhances the visibility of the contact surface contour in the embedded camera view.

\begin{figure}[htbp]
  \centering
  \includegraphics[width=\linewidth]{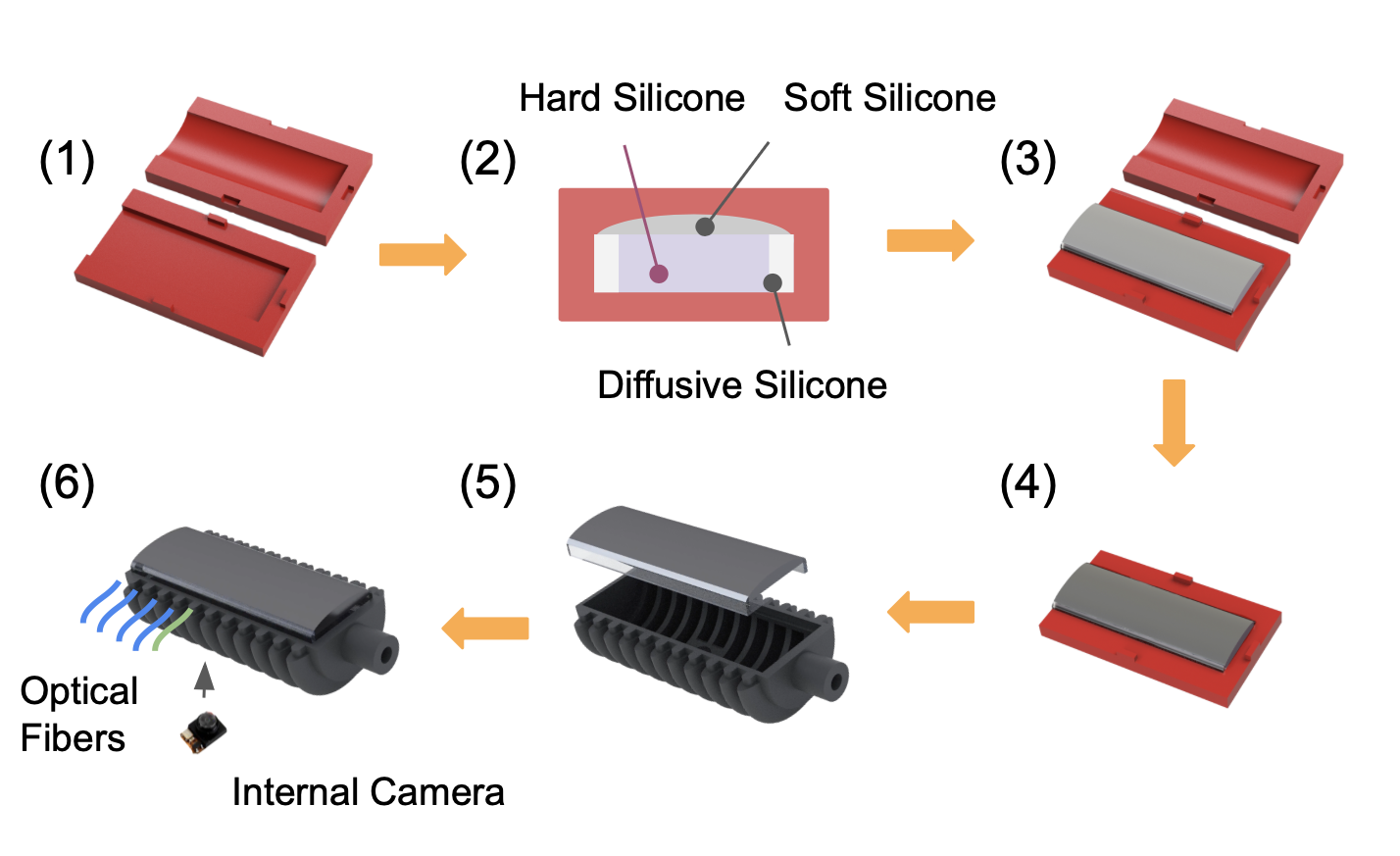}
  \caption[A, B]{A, Fabrication process of proposed PneuGelSight. (1) Assembled mold (2) Silicone resin is poured into the mold. (3) After curing, the top mold is removed. (4) Aluminum powder is applied as semi-specular coating. (5) 3D printed body is connected with the composite silicone slab. (6) Optical fiber and internal camera are sealed to the body. (7) Relaxed stage of the gripper. B, Unactuated robot configuration with dimensions.
 }
  \label{fig: fabrication}
\end{figure}

\section{Optical Simulation}

\subsection{Optical Models}
To ensure accurate simulation results, it is necessary to model the optical properties of all components of the soft robot gripper that can interact with light. This includes the optical fibers, diffusive elastomer layer, transparent silicone gel layer, reflective coating layer, and camera. However, the soft robot body will not be modeled, as we have painted its surface black, which absorbs all light from the inside and blocks all light from the outside environment. In the following paragraphs, we will discuss each part in detail.

Optical fibers are waveguides that can transport electromagnetic waves through them with little to no energy loss. The electromagnetic wave profiles allowed to transport through optical fibers are called guided modes and are determined by the diameter and material of the optical fibers\cite{waveguides}. The optical fibers used in our robot have relatively large diameters (0.75mm in diameter) which allows a large number of guided modes so that the full energy profile could be transported. Consequently, the distribution of the outgoing radiance of the optical fibers entirely relies on the distribution of radiance of the source LEDs. 

To minimize the sim-to-real gap, we simulate optical fibers and diffusive layers together as a set of point lights. Figure \ref{fig: optimize sensor}A shows a comparison: the left column displays real-world images captured from the embedded camera while the right column shows simulated images under similar bending angles. With the same color arrangement and bending angles, the images in different domains share the same pattern, which proves the effectiveness of our proposed PBR pipeline.

The light models of other materials are straightforward. We use the rough dielectric model to model the transparent silicone layer because the rough dielectric model is suitable for a homogeneous transparent material with a slightly uneven surface. The coating layer is modeled by a surface diffusive model in the previous work by our lab\cite{agarwal2021simulationvisionbasedtactilesensors}. The wide-angle camera can be directly modeled as a perspective camera with a certain field of view and resolution. We neglect the geometry distortion because we will preprocess all the images with the de-warping method for fisheye cameras provided by Opencv library during actual normal prediction.

\subsection{Design Space Exploration}

Now that we have our soft robot gripper modeled in the simulation and the color variance of the indented part as a metric to describe the sensing performance (mentioned in the main paper), we can proceed to the optimization stage. In the experiment, we observe that changes in the light source direction do not have as much effect on sensing performance as color choice because of our strong diffusion layer. Therefore, we have 24 color choices of the light source to optimize. 

\begin{figure*}[htbp]
  \centering
  \includegraphics[width=\linewidth]{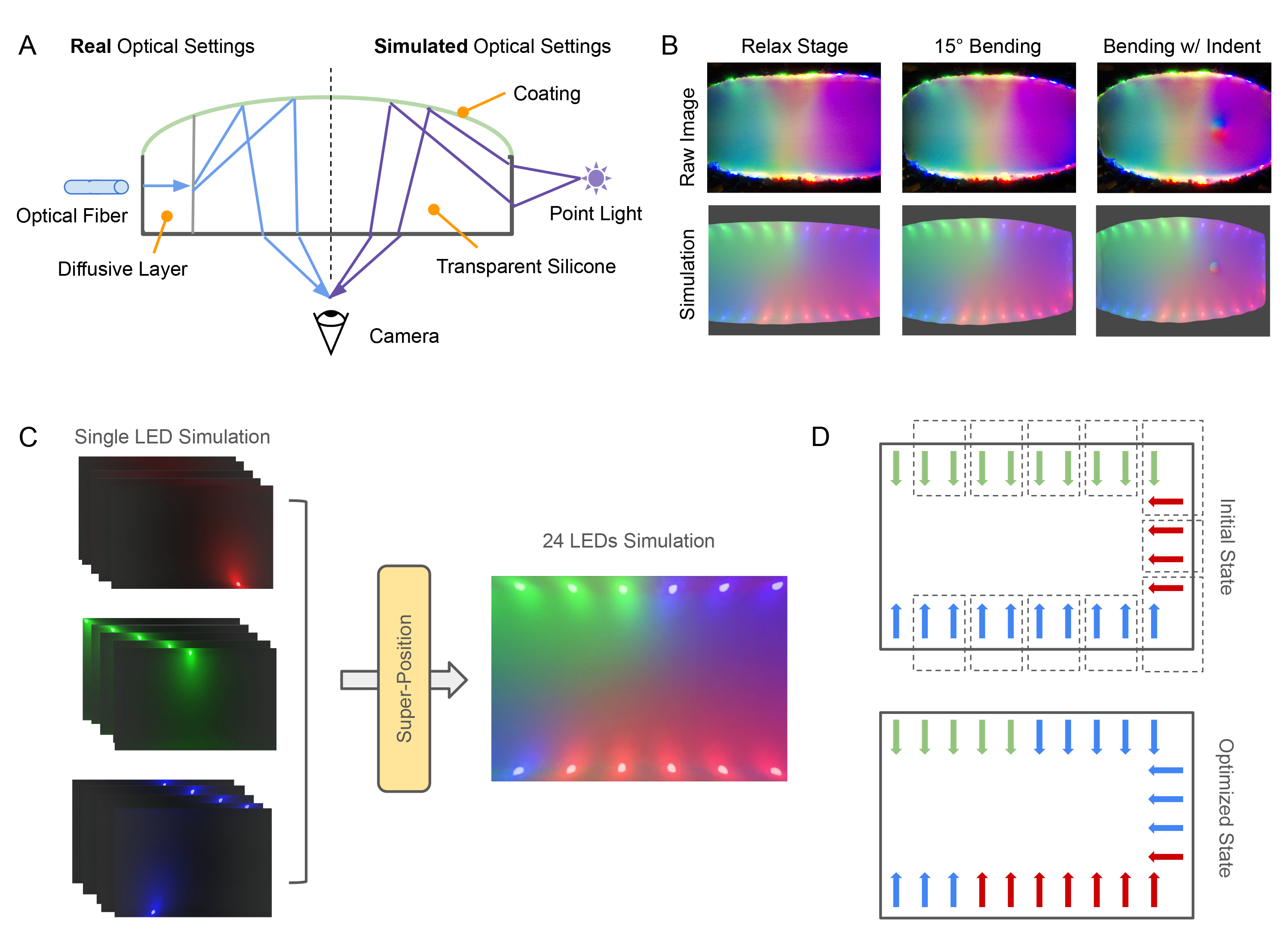}
  \caption[A, B, C]{\textbf{(A)}, Comparison between real-world and simulated image settings. \textbf{(B)}, Examples of real and simulated images across various scenarios. \textbf{(C)}, Visualization of the superposition of multiple light sources. A total of 72 images (3 sets of 24) were rendered, followed by a grid search based on color variance metrics. \textbf{(D)}, Initial and final states before and after optimization, with the applied grid layout indicated by the dotted box in the figure showing the initial state.}
  \label{fig: optical simulation}
\end{figure*}

We do a grid search to iterate all possible combinations of color choices to find out the best color choice. Here, we use two tricks to reduce the work. Leveraging the superposition of light, we render each simulation with only one LED on. While testing on a specific color choice, we superpose the rendering result by adding 24 images with the correct color layout. By doing so, we reduce the number of renderings from $3^{24}$ down to 3 × 24. Also, we notice that groups of neighboring several optical fibers with the same colored light source tend to have a higher metric value than all neighboring optical fibers that have different colored light sources. Therefore, we propose to group the two nearby optical fibers with the same color light source. We visualize the initial and optimized states in Figure \ref{fig: optical simulation}C, with bounding boxes indicating the grouped optical fibers.

\end{document}